\documentclass[11pt]{article}
\usepackage{acl2016}
\usepackage{times}
\usepackage{url}
\usepackage{latexsym}
\usepackage{graphicx}
\usepackage{algorithm}
\usepackage{algorithmic}
 \usepackage{verbatim}
\usepackage{amsfonts}
\usepackage{multirow}
\usepackage{amssymb}
\usepackage{color}
\usepackage{enumitem}
\usepackage{microtype}
\usepackage{booktabs}
\usepackage{rotating}
\usepackage{balance}

\makeatletter
\let\@fnsymbol\@alph
\newcommand{\@BIBLABEL}{\@emptybiblabel}
\newcommand{\@emptybiblabel}[1]{}
\makeatother
\usepackage{hyperref}
\hypersetup{
    colorlinks,
    linkcolor={blue!25!black},
    citecolor={blue!25!black},
    urlcolor={blue!25!black}
}

\aclfinalcopy

\usepackage{url} 
\usepackage{cleveref}
\Crefformat{section}{\S#2#1#3}
\Crefformat{equation}{(#2#1#3)}

\usepackage{xargs} 
\usepackage[colorinlistoftodos,prependcaption,textsize=tiny]{todonotes}
\usepackage{marginnote}
\newcommandx{\todoir}[2][1=]{\todo[inline,backgroundcolor=red!10]{SR: #2}\xspace}
\newcommandx{\todoix}[2][1=]{\todo[inline,backgroundcolor=green!10]{KX: #2}\xspace}
\newcommandx{\todoif}[2][1=]{\todo[inline,backgroundcolor=yellow!10]{YF: #2}\xspace}

\newcommandx{\todor}[2][1=]{\todo[linecolor=red,backgroundcolor=red!10,bordercolor=red,#1]{SR: #2}\xspace}
\newcommandx{\todox}[2][1=]{\todo[linecolor=green,backgroundcolor=green!10,bordercolor=green,#1]{KX: #2}\xspace}
\newcommandx{\todof}[2][1=]{\todo[linecolor=yellow,backgroundcolor=yellow!10,bordercolor=yellow,#1]{YF: #2}\xspace}

\def\aclpaperid{34} 

\usepackage{xspace}
\newcommand \webq{WebQuestions\xspace}

\title{Question Answering on Freebase via Relation Extraction and Textual~Evidence}

 \author{Kun Xu$^{1}$ \and Siva Reddy$^{2}$ \and Yansong Feng{${^{1,}}\thanks{\quad Corresponding author}$} \and Songfang Huang$^{3}$ \and Dongyan Zhao$^{1}$ \\
         $^{1}$Institute of Computer Science \& Technology, Peking University, Beijing, China \\ 
         $^{2}$School of Informatics, University of Edinburgh, UK \\
	 $^{3}$IBM China Research Lab, Beijing, China \\
	\{\texttt{xukun}, \texttt{fengyansong}, \texttt{zhaody}\}\texttt{@pku.edu.cn} \\
	\texttt{siva.reddy@ed.ac.uk} \\
	\texttt{huangsf@cn.ibm.com} \\
}

\date{}

\begin{document}

\maketitle

\begin{abstract}
Existing knowledge-based question answering systems often rely on small annotated training data. While shallow methods like relation extraction are robust to data scarcity, they are less expressive than the deep meaning representation methods like semantic parsing, thereby failing at answering questions involving multiple constraints. Here we alleviate this problem by empowering a relation extraction method with additional evidence from Wikipedia. We first present a neural network based relation extractor to retrieve the candidate answers from Freebase, and then infer over Wikipedia to validate these answers. Experiments on the WebQuestions question answering dataset show that our method achieves an $F_1$ of 53.3\%, a substantial improvement over the state-of-the-art. 
\end{abstract}

\section{Introduction}
Since the advent of large structured knowledge bases (KBs) like Freebase \cite{DBLP:dblp_conf/sigmod/BollackerEPST08}, 
YAGO \cite{DBLP:conf/www/SuchanekKW07} and DBpedia 
\cite{DBLP:dblp_conf/semweb/AuerBKLCI07}, answering natural language questions using those structured KBs, also known as KB-based question answering (or KB-QA), is attracting increasing research efforts from both natural language processing and information retrieval communities.

The state-of-the-art methods for this task can be roughly categorized into two streams. The first is based on semantic parsing 
\cite{DBLP:conf/emnlp/BerantCFL13,DBLP:conf/emnlp/KwiatkowskiCAZ13}, which typically learns a grammar that can parse natural language 
to a sophisticated meaning representation language. But such sophistication requires a lot of annotated training examples that contains compositional 
structures, a practically impossible solution for large KBs such as Freebase. Furthermore, mismatches between grammar predicted structures and KB structure is also a common problem \cite{DBLP:conf/emnlp/KwiatkowskiCAZ13,DBLP:conf/acl/BerantL14,reddy14}. 

On the other hand, instead of building a formal meaning representation, information extraction methods retrieve a set of candidate answers from KB using relation extraction
\cite{yao-jacana-freebase-acl2014,msr14,yao-scratch-qa-naacl2015,DBLP:conf/cikm/BastH15} or distributed representations \cite{D14-1067,dong-EtAl:2015:ACL-IJCNLP1}. Designing large training datasets for these methods is relatively easy \cite{yao-jacana-freebase-acl2014,DBLP:journals/corr/BordesUCW15,serban:2016}. These methods are often good at producing an answer irrespective of their correctness. However, handling compositional questions that involve multiple entities and relations, still remains a challenge. Consider the question \textsl{what mountain is the highest in north america}. Relation extraction methods typically answer with all the mountains in \textsl{North America} because of the lack of sophisticated representation for the mathematical function \textsl{highest}. To select the correct answer, one has to retrieve all the heights of the mountains, and sort them in descending order, and then pick the first entry. We propose a method based on textual evidence which can answer such questions without solving the mathematic functions implicitly.

Knowledge bases like Freebase capture real world facts, and Web resources like Wikipedia provide a large repository of sentences that validate or support these facts. For example, a sentence in Wikipedia says, \textsl{Denali (also known as Mount McKinley, its former official name) is the highest mountain peak in North America, with a summit elevation of 20,310 feet (6,190 m) above sea level}. To answer our example question against a KB using a relation extractor, we can use this sentence as external evidence, filter out wrong answers and pick the correct one. 

Using textual evidence not only mitigates representational issues in relation extraction, but also alleviates the data scarcity problem to some extent. Consider the question, \textsl{who was queen isabella's mother}. Answering this question involves predicting two constraints hidden in the word \textsl{mother}. One constraint is that the answer should be the \textsl{parent} of \textsl{Isabella}, and the other is that the answer's \textsl{gender} is \textsl{female}. Such words with multiple latent constraints have been a pain-in-the-neck for both semantic parsing and relation extraction, and requires larger training data (this phenomenon is coined as sub-lexical compositionality by \newcite{wang2015}). Most systems are good at triggering the \textsl{parent} constraint, but fail on the other, i.e., the answer entity should be \textit{female}. Whereas the textual evidence from Wikipedia, \textsl{\ldots her mother was Isabella of Barcelos \ldots}, can act as a further constraint to answer the question correctly.

We present a novel method for question answering which infers on both structured and unstructured resources. Our method consists of two main steps as outlined in~\Cref{sec:overview}. In the first step we extract answers for a given question using a structured KB (here Freebase) by jointly performing entity linking and relation extraction~(\Cref{sec:kb-qa}). In the next step we validate these answers using an unstructured resource (here Wikipedia) to prune out the wrong answers and select the correct ones~(\Cref{sec:refine}). Our evaluation results on a benchmark dataset \webq show that our method outperforms existing state-of-the-art models. Details of our experimental setup and results are presented in~\Cref{sec:experiments}. Our code, data and results can be downloaded from {\small \url{https://github.com/syxu828/QuestionAnsweringOverFB}}.

\begin{figure}[t]
\centering\includegraphics[width=0.5\textwidth]{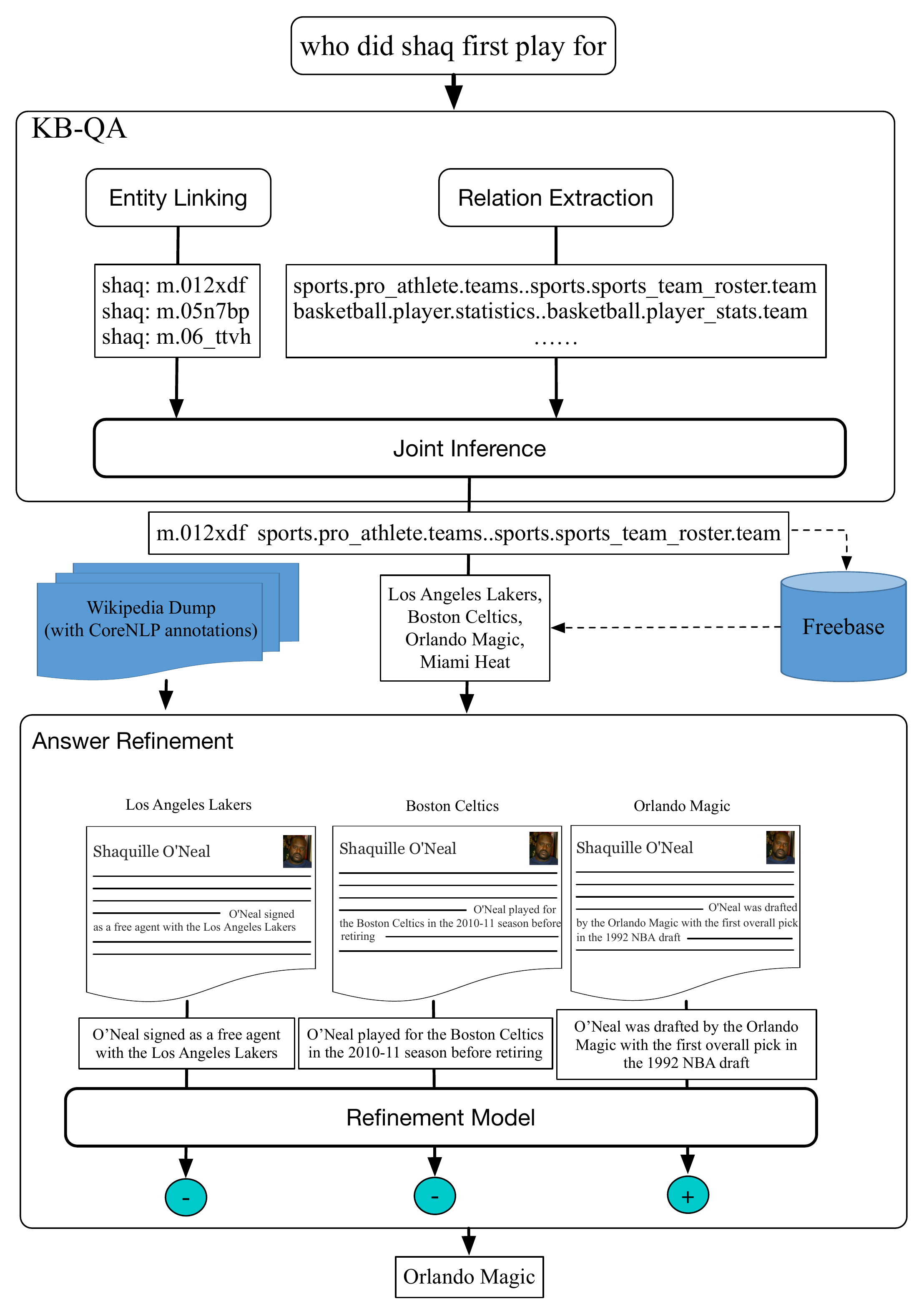} 
\caption{An illustration of our method to find answers for the given question \textit{who did shaq first play for}.}
\label{fig:qaframework}
\end{figure}

\section{Our Method}
\label{sec:overview}

\Cref{fig:qaframework} gives an overview of our method for the question ``\textsl{who did shaq first play for}''. We have two main steps: (1) inference on Freebase (\textit{KB-QA} box); and (2) further inference on Wikipedia (\textit{Answer Refinement} box). Let us take a close look into step~1. Here we perform \textit{entity linking} to identify a topic entity in the question and its possible Freebase entities. We employ a \textit{relation extractor} to predict the potential Freebase relations that could exist between the entities in the question and the answer entities. Later we perform a \textit{joint inference} step over the entity linking and relation extraction results to find the best entity-relation configuration which will produce a list of candidate answer entities. In the step~2, we refine these candidate answers by applying an \textit{answer refinement} model which takes the Wikipedia page of the topic entity into consideration to filter out the wrong answers and pick the correct ones. 

While the overview in~\Cref{fig:qaframework} works for questions containing single Freebase relation, it also works for questions involving multiple Freebase relations. Consider the question \textsl{who plays anakin skywalker in star wars 1}. The actors who are the answers to this question should satisfy the following constraints: (1) the actor played \textit{anakin skywalker}; and (2) the actor played in \textsl{star wars~1}. Inspired by \newcite{msra14}, we design a dependency tree-based method to handle such multi-relational questions. We first decompose the original question into a set of \textit{\mbox{sub-questions}} using syntactic patterns which are listed in Appendix. The final answer set of the original question is obtained by intersecting the answer sets of all its sub-questions. For the example question, the sub-questions are \textit{who plays anakin skywalker} and \textsl{who plays in star wars~1}. These sub-questions are answered separately over Freebase and Wikipedia, and the intersection of their answers to these sub-questions is treated as the final answer.




\section{Inference on Freebase}
\label{sec:kb-qa}

Given a sub-question, we assume the question word\footnote{who, when, what, where, how, which, why, whom, whose.} that represents the answer has a distinct KB relation~$r$ with an entity $e$ found in the question, and predict a single KB triple $(e,r,?)$ for each sub-question (here $?$ stands for the answer entities). The QA problem is thus formulated as an information extraction problem that involves two sub-tasks, i.e., entity linking and relation extraction. We first introduce these two components, and then present a joint inference procedure which further boosts the overall performance.

\subsection{Entity Linking}
For each question, we use hand-built sequences of part-of-speech categories to identify all possible named entity mention spans, e.g., the sequence \textit{NN} (\textit{shaq}) may indicate an entity. For each mention span, we use the entity linking tool S-MART\footnote{\small S-MART demo can be accessed at \\{\small\url{http://msre2edemo.azurewebsites.net/}}} \cite{yang-chang:2015:ACL-IJCNLP} to retrieve the top 5~entities from Freebase. These entities are treated as candidate entities that will eventually be disambiguated in the joint inference step. For a given mention span, S-MART first retrieves all possible entities of Freebase by surface matching, and then ranks them using a statistical model, which is trained on the frequency counts with which the surface form occurs with the entity. 

\subsection{Relation Extraction\label{MCCNN}}
We now proceed to identify the relation between the answer and the entity in the question. Inspired by the recent success of neural network models in KB question-answering \cite{yih-EtAl:2015:ACL-IJCNLP,dong-EtAl:2015:ACL-IJCNLP1}, and the success of syntactic dependencies for relation extraction \cite{liu-EtAl:2015:ACL-IJCNLP,xu-EtAl:2015:EMNLP1}, we propose a Multi-Channel Convolutional Neural Network (MCCNN) which could exploit both syntactic and sentential information for relation extraction. 


\begin{figure}[t]
\centering\includegraphics[width=0.49\textwidth]{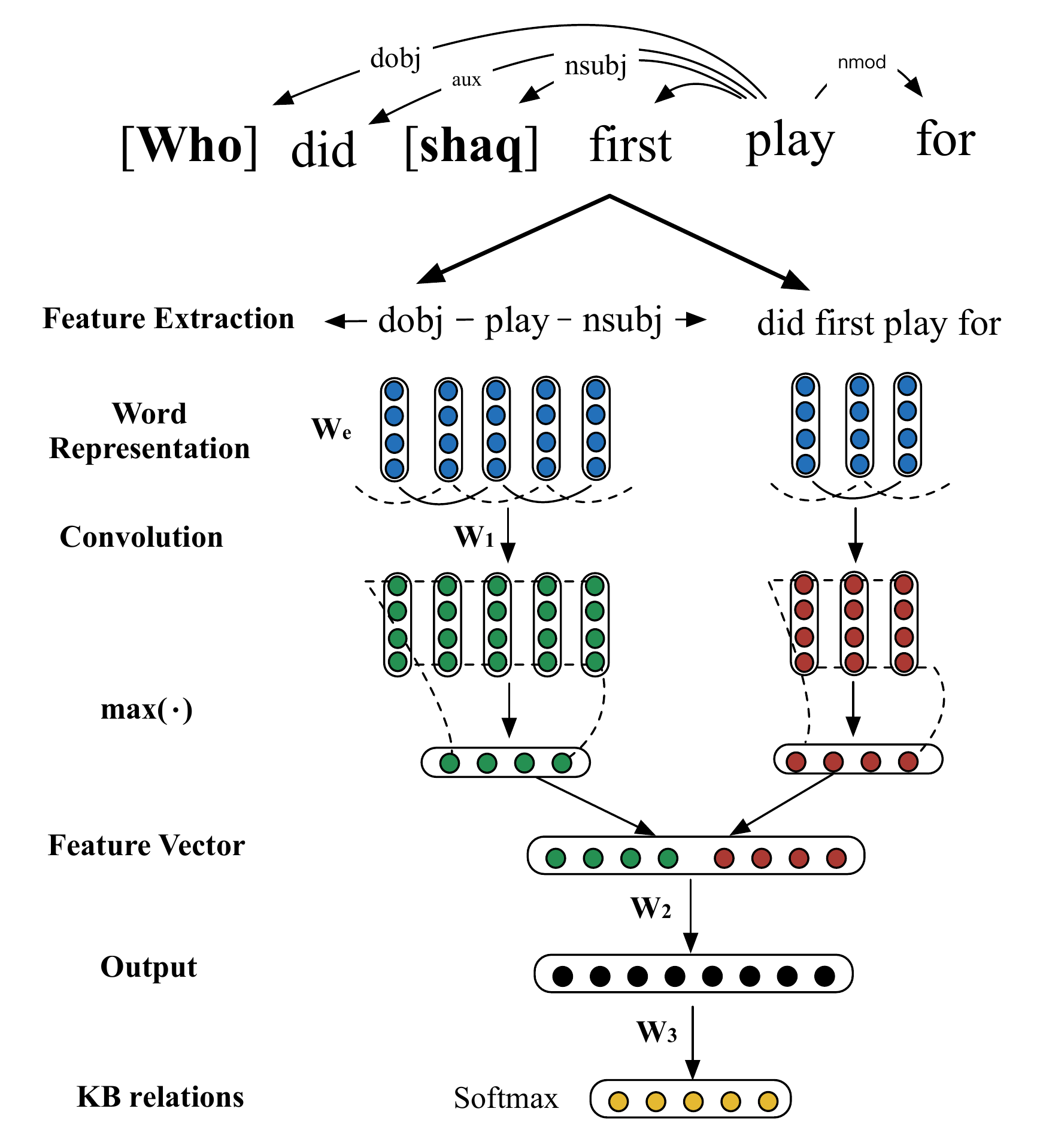} 
\caption{ \label{cnn_architecture}Overview of the multi-channel convolutional neural network for relation extraction.
$W_e$ is the word embedding matrix, W$_1$ is the convolution matrix, W$_2$ is the activation matrix and W$_3$ is the classification matrix.
}
\vspace{-0.3cm}
\end{figure}

\subsubsection{MCCNNs for Relation Classification}
In MCCNN, we use two channels, one for syntactic information and the other for sentential information.  The network structure is illustrated in Figure~\ref{cnn_architecture}. Convolution layer tackles an input of varying length returning a fixed length vector (we use max pooling) for each channel. These fixed length vectors are concatenated and then fed into a \textit{softmax} classifier, the output dimension of which is equal to the number of predefined relation types. The value of each dimension indicates the confidence score of the corresponding relation. 

\paragraph{Syntactic Features}
We use the shortest path between an entity mention and 
the question word in the dependency tree\footnote{We use Stanford CoreNLP 
dependency parser \cite{manning-EtAl:2014:P14-5}.} as input to the first channel. Similar to \newcite{xu-EtAl:2015:EMNLP1}, we treat the path as a 
concatenation of vectors of words, dependency edge directions and 
dependency labels, and feed it to the convolution layer.
Note that, the entity mention and the question word are 
excluded from the dependency path so as to learn a more general relation representation 
in syntactic level. As shown in Figure~\ref{cnn_architecture}, 
the dependency path between \textsl{who} and \textsl{shaq} is 
$\leftarrow$\textsl{~dobj~--~play~--~nsubj~}$\rightarrow$.

\paragraph{Sentential Features}
This channel takes the words in the sentence as input excluding the question word and the entity mention. As illustrated in 
Figure~\ref{cnn_architecture}, the vectors for \textsl{did, first, play} and \textsl{for} are fed into this channel.

\subsubsection{Objective Function and Learning}
The model is learned using pairs of question and its corresponding gold relation from the training data. Given an input question 
$x$ with an annotated entity mention, the network outputs a vector $o(x)$, where 
the entry $o_{k}(x)$ is the probability that there exists the \textit{k}-th 
relation between the entity and the expected answer. We denote $t(x) \in 
\mathbb{R}^{K\times 1}$ as the target distribution vector, in which the value 
for the gold relation is set to~1, and others to~0. We compute the cross entropy error 
between $t(x)$ and $o(x)$, and further define the objective function over the training data as:
\begin{displaymath}
  J(\theta) = - \sum_{x} \sum_{k=1}^{K} t_k(x) \log o_k(x) + \lambda||\theta||^{2}_{2} 
\end{displaymath}
where $\theta$ represents the weights, and $\lambda$ the $L2$~regularization 
parameters. The weights $\theta$ can be efficiently computed via 
back-propagation through network structures. To minimize $J(\theta)$, we apply 
stochastic gradient descent (SGD) with AdaGrad 
\cite{DBLP:journals/jmlr/DuchiHS11}.

\subsection{\normalsize Joint~Entity~Linking~\&~Relation~Extraction\label{sec:jointInference}}
A pipeline of entity linking and relation extraction may suffer from error propagations.
As we know, entities and relations have strong selectional preferences that 
certain entities do not appear with certain relations and vice versa. Locally optimized models could not exploit these implicit \textit{bi-directional} preferences.
Therefore, we use a joint model to find a globally optimal 
entity-relation assignment from local predictions.
The key idea behind is to leverage various clues from the two local models
and the KB to rank a correct entity-relation assignment higher than other combinations. We describe the learning procedure and the features below.

\subsubsection{Learning}
Suppose the pair $(e_{gold}, r_{gold})$ represents the gold entity/relation pair for a question $q$. We take all our entity and relation predictions for $q$, create a list of entity and relation pairs $\{(e_{0}, r_{0}), (e_{1}, r_{1}), ..., (e_{n}, r_{n})\}$ from $q$ and rank them using an \textsc{svm} rank classifier \cite{DBLP:conf/kdd/Joachims06} which is trained to predict a rank for each pair. Ideally higher rank indicates the prediction is closer to the gold prediction. For training, \textsc{svm} rank classifier requires a ranked or scored list of entity-relation pairs as input. We create the training data containing ranked input pairs as follows: if both $e_{pred} = e_{gold}$ and $r_{pred} = r_{gold}$, we assign it with a score of~3. If only the entity or relation equals to the gold one (i.e., $e_{pred}=e_{gold}$, $r_{pred}\neq r_{gold}$ or $e_{pred}\neq e_{gold}$, $r_{pred}=r_{gold}$), we assign a score of~2 (encouraging partial overlap). When both entity and relation assignments are wrong, we assign a score of~1.

\subsubsection{Features}
For a given entity-relation pair, we extract the following features which are passed as an input vector to the \textsc{svm} ranker above:

\vspace{-0.2cm}
\paragraph{Entity Clues.}
We use the score of the predicted entity returned by the entity linking system as a feature. 
The number of word overlaps between the entity mention and entity's Freebase name is also included as a feature. 
In Freebase, most entities have a relation \textit{fb:description} which describes the entity. 
For instance, in the running example, \textit{shaq} is linked to three potential entities \textsl{m.06\_ttvh (Shaq Vs. Television Show)},
\textsl{m.05n7bp (Shaq Fu Video Game)} and \textsl{m.012xdf (Shaquille O'Neal)}.
Interestingly, the word \textit{play} only appears in the description of \textsl{Shaquille O'Neal} and it occurs three times. We count the content word overlap between the given question and the entity's description, and include it as a feature.

\vspace{-0.2cm}
\paragraph{Relation Clues.} The score of relation returned by the MCCNNs 
is used as a feature. Furthermore, we view each relation as a \textit{document} which consists of 
the training questions that this relation is expressed in. For a given 
question, we use the sum of the \textit{tf-idf} scores of its words with respect to the relation as a feature. A Freebase relation~$r$ is a concatenation of a series of fragments $r~=~r_1.r_2.r_3$. For instance, the three fragments of 
\textsl{people.person.parents} are \textsl{people}, \textsl{person} and 
\textsl{parents}. The first two fragments indicate the Freebase type of the
subject of this relation, and the third fragment indicates the object 
type, in our case the answer type. We use an indicator feature to denote if the surface form of the third fragment (here \textsl{parents}) appears in the question. 

\vspace{-0.2cm}
\paragraph{Answer Clues.}
The above two feature classes indicate local features. From the entity-relation $(e,r)$ pair, we create the query triple $(e,r,?)$ to retrieve the answers, and further extract features from the answers. These features are non-local since we require both~$e$ and~$r$ to retrieve the answer. One such feature is using the co-occurrence of the answer type and the question word based on the intuition that question words often indicate the answer type, e.g., the question word \textsl{when} usually indicates the answer type \textsl{type.datetime}. Another feature is the number of answer entities retrieved. 

\section{Inference on Wikipedia\label{sec:refine}}
We use the best ranked entity-relation pair from the above step to retrieve candidate answers from Freebase. 
In this step, we validate these answers using Wikipedia as our unstructured knowledge resource where most statements in it are verified for factuality by multiple people.

Our refinement model is inspired by the intuition of how people refine their answers. 
If you ask someone: \textsl{who did shaq first play for},
and give them four candidate answers (\textsl{Los Angeles Lakers}, \textsl{Boston Celtics}, \textsl{Orlando Magic} and \textsl{Miami Heat}),
as well as access to Wikipedia, that person might 
first determine that the question is about \textsl{Shaquille O'Neal}, then go to \textsl{O'Neal}'s Wikipedia page, and search for the
sentences that contain the candidate answers as evidence.
By analyzing these sentences, one can figure out whether a candidate answer is correct or not.

\subsection{Finding Evidence from Wikipedia}
As mentioned above, we should first find the Wikipedia page corresponding to the \textit{topic entity} in the given question. We use Freebase API to convert Freebase entity to Wikipedia page. We extract the content from the Wikipedia page and process it with \textit{Wikifier} \cite{DBLP:conf/emnlp/ChengR13} which recognizes Wikipedia entities, which can further be linked to Freebase entities using Freebase API. Additionally we use Stanford CoreNLP \cite{manning-EtAl:2014:P14-5} for tokenization and entity co-reference resolution. We search for the sentences containing the candidate answer entities retrieved from Freebase. 
For example, the Wikipedia page of \textit{O'Neal} contains a sentence ``\textsl{O'Neal was drafted by the Orlando
Magic with the first overall pick in the 1992 NBA draft}'', which is taken into account by the \textit{refinement model} (our inference model on Wikipedia) to discriminate whether \textsl{Orlando Magic} is the answer for the given question.

\subsection{Refinement Model}
We treat the refinement process as a binary classification task over the candidate answers, i.e., correct (positive) and incorrect (negative) answer. 
We prepare the training data for the refinement model as follows.
On the training dataset, we first infer on Freebase to retrieve the candidate answers. 
Then we use the annotated gold answers of these questions and Wikipedia to create the training data. 
Specifically, we treat the sentences that contain correct/incorrect answers as positive/negative examples for the refinement model.
We use \textsc{libsvm} \cite{DBLP:journals/tist/ChangL11} to learn the weights for classification.

Note that, in the Wikipedia page of the topic entity,
we may collect more than one sentence that contain a candidate answer.
However, not all sentences are relevant, therefore we consider the candidate answer as correct \textit{if} at least there is one positive evidence.
On the other hand, sometimes, we may not find any evidence for the candidate answer.
In these cases, we fall back to the results of the KB-based approach.

\subsection{Lexical Features}
Regarding the features used in \textsc{libsvm}, we use the following lexical features extracted from the question and a Wikipedia sentence.
Formally, given a question $q$~=~$<$$q_1$,~...~$q_{n}$$>$ and an evidence sentence $s$~=~$<$$s_1$,~...~$s_{m}$$>$, 
we denote the tokens of $q$ and~$s$ by $q_i$ and $s_j$, respectively.
For each pair ($q$,~$s$), we identify a set of all possible token pairs ($q_i$,~$s_j$), the occurrences of which are used as features. 
As learning proceeds, we hope to learn a higher weight for a feature like 
(\textsl{first},~\textsl{drafted}) and a lower weight for (\textsl{first},~\textsl{played}).

\section{Experiments\label{sec:experiments}}
In this section we introduce the experimental setup, the main results and detailed analysis of our system.

\subsection{Training and Evaluation Data}
We use the \webq \cite{DBLP:conf/emnlp/BerantCFL13} dataset, which contains 5,810 questions crawled via Google Suggest service, with answers annotated on Amazon Mechanical Turk. 
The questions are split into training and test sets, which contain 3,778 questions (65\%) and 2,032 questions (35\%), respectively. 
We further split the training questions into 80\%/20\% for development.

To train the MCCNNs and the joint inference model, we need the gold standard relations of the questions. 
Since this dataset contains only question-answer pairs and annotated topic entities, instead of relying on gold relations we rely on \textit{surrogate} gold relations which produce answers that have the highest overlap with gold answers. 
Specifically, for a given question, we first locate the topic entity $e$ in the Freebase graph, then select 1-hop and 2-hop relations connected to the topic entity as relation candidates. 
The 2-hop relations refer to the $n$-ary relations of Freebase, i.e., first hop from the subject to a mediator node, and the second from the mediator to the object node.
For each relation candidate~$r$, we issue the query ($e$,~$r$,~$?$) to the KB, and label the relation that produces the answer with minimal \mbox{$F_1$-loss} against the gold answer, as the \textit{surrogate} gold relation.
From the training set, we collect 461~relations to train the MCCNN, and the target prediction during testing time is over these relations.

\subsection{Experimental Settings}
We have 6~dependency tree patterns based on \newcite{msra14} to decompose the question into sub-questions (See Appendix). We initialize the word embeddings with \newcite{DBLP:conf/acl/TurianRB10}'s word representations with dimensions set to~50. 
The hyper parameters in our model are tuned using the development set. The window size of MCCNN is set to~3. The sizes of the hidden layer~1 and the hidden layer~2 of the two MCCNN channels are set to 200 and 100, respectively.
We use the Freebase version of \newcite{DBLP:conf/emnlp/BerantCFL13}, containing 4M entities and 5,323 relations.

\begin{table}[t]
\small
\centering
\begin{tabular}{p{5.5cm} c}
\toprule
Method &average $F_1$\\
\midrule
\newcite{DBLP:conf/emnlp/BerantCFL13} & 35.7 \\
\newcite{yao-jacana-freebase-acl2014} & 33.0 \\
\newcite{DBLP:conf/nlpcc/XuZFZ14} & 39.1 \\
\newcite{DBLP:conf/acl/BerantL14} & 39.9 \\
\newcite{msra14} & 37.5 \\
\newcite{D14-1067} & 39.2 \\
\newcite{dong-EtAl:2015:ACL-IJCNLP1} & 40.8 \\
\newcite{yao-scratch-qa-naacl2015} & 44.3 \\
\newcite{DBLP:conf/cikm/BastH15} & 49.4 \\
\newcite{berant_imitation_2015} & 49.7 \\
\newcite{reddy_transforming_2016} & 50.3 \\
\newcite{yih-EtAl:2015:ACL-IJCNLP} & 52.5 \\
\midrule
\multicolumn{2}{c}{This work} \\
\midrule
Structured &  44.1 \\
Structured + Joint & 47.1 \\
Structured + Unstructured & 47.0 \\
Structured + Joint + Unstructured & \textbf{53.3} \\
\bottomrule
\end{tabular}
\caption{Results on the test set.}
\label{tab:results}
\end{table}

\subsection{Results and Discussion}
We use the average question-wise~$F_1$ as our evaluation metric.\footnote{We use the evaluation script available at \url{http://www-nlp.stanford.edu/software/sempre}.} To give an idea of the impact of different configurations of our method, we compare the following with existing methods.

\paragraph{Structured.} This method involves inference on Freebase only. First the entity linking~(EL) system is run to predict the topic entity. Then we run the relation extraction~(RE) system and select the best relation that can occur with the topic entity. We choose this entity-relation pair to predict the answer.

\paragraph{Structured + Joint.} In this method instead of the above pipeline, we perform joint EL and RE as described in \Cref{sec:jointInference}. 

\paragraph{Structured+Unstructured.} We use the pipelined EL and RE along with inference on Wikipedia as described in \Cref{sec:refine}.

\paragraph{Structured + Joint + Unstructured.} This is our main model. We perform inference on Freebase using joint EL and RE, and then inference on Wikipedia to validate the results.
Specifically, we treat the top two predictions of the joint inference model as the candidate subject and relation pairs, and extract the corresponding answers from each pair, take the union, and filter the answer set using Wikipedia.


Table~\ref{tab:results} summarizes the results on the test data along with the results from the literature.\footnote{We use development data for all our ablation experiments. Similar trends are observed on both development and test results.} We can see that joint EL and RE performs better than the default pipelined approach, and outperforms most semantic parsing based models, except \cite{berant_imitation_2015} which searches partial logical forms in strategic order by combining imitation learning and agenda-based parsing. In addition, inference on unstructured data helps the default model. The joint EL and RE combined with inference on unstructured data further improves the default pipelined model by 9.2\% (from~44.1\% to~53.3\%), and achieves a new state-of-the-art result beating the previous reported best result of \newcite{yih-EtAl:2015:ACL-IJCNLP} (with one-tailed t-test significance of~$p < 0.05$).


\begin{table}[!t]
\small
\centering
\begin{tabular}{ccc}
\toprule
 & Entity Linking & Relation Extraction \\
 & Accuracy & Accuracy \\
\midrule
Isolated Model & 79.8 & 45.9  \\
Joint Inference  & \textbf{83.2}  & \textbf{55.3} \\ 
\bottomrule
\end{tabular}
\caption{Impact of the joint inference on the development set}
\label{tab:ELandREresults}
\end{table}

\begin{table}[t]
\small
\centering
\begin{tabular}{p{5.1cm} c}
\toprule
Method & average $F_1$\\
\midrule
Structured (syntactic) & 38.1 \\
Structured (sentential) &  38.7 \\
Structured (syntactic + sentential) & 40.1 \\
\midrule
Structured + Joint (syntactic) & 43.6 \\
Structured + Joint (sentential) & 44.1 \\
Structured + Joint (syntactic + sentential) & \textbf{45.8} \\
\bottomrule
\end{tabular}
\caption{Impact of different MCCNN channels on the development set.}
\label{tab:ResultsOnDev}
\vspace{-0.3cm}
\end{table}

\subsubsection{Impact of Joint EL \& RE}
From Table~\ref{tab:results}, we can see that the joint EL \& RE gives a performance boost of 3\% (from 44.1 to 47.1). We also analyze the impact of joint inference on the individual components of EL \& RE. 

We first evaluate the EL component using the gold entity annotations on the development set. As shown in Table~\ref{tab:ELandREresults}, for 79.8\% questions, our entity linker can correctly find the gold standard topic entities. The joint inference improves this result to 83.2\%, a 3.4\% improvement. Next we use the \textit{surrogate} gold relations to evaluate the performance of the RE component on the development set. As shown in Table~\ref{tab:ELandREresults}, the relation prediction accuracy increases by 9.4\% (from 45.9\% to 55.3\%) when using the joint inference.

\subsubsection{Impact of the Syntactic and the Sentential Channels}
Table~\ref{tab:ResultsOnDev} presents the results on the impact of individual and joint channels on the end QA performance. When using a single-channel network, we tune the parameters of only one channel while switching off the other channel. 
As seen, the sentential features are found to be more important than syntactic features. We attribute this to the short and noisy nature of \webq questions due to which syntactic parser wrongly parses or the shortest dependency path does not contain sufficient information to predict a relation.
By using both the channels, we see further improvements than using any one of the channels.

\begin{table}[t]
\small
\centering
\begin{tabular}{|l|}
\hline
\multicolumn{1}{|c|}{Question \& Answers} \\
\hline
\hline
$1$. what is the largest nation in europe \\ 
Before: \textcolor{red}{Kazakhstan}, \textcolor{red}{Turkey}, \textcolor{blue}{Russia}, ... \\
After: \textcolor{blue}{Russia} \\ \cline{1-1}
$2$. which country in europe has the largest land area \\
Before: \textcolor{red}{Georgia}, \textcolor{red}{France}, \textcolor{blue}{Russia}, ... \\
After: \textcolor{red}{Russian Empire}, \textcolor{blue}{Russia}  \\ \cline{1-1}
$3$. what year did ray allen join the nba  \\
Before: \textcolor{red}{2007}, \textcolor{red}{2003}, \textcolor{blue}{1996}, \textcolor{red}{1993}, \textcolor{red}{2012}  \\
After: \textcolor{blue}{1996}  \\
\hline
\hline
$4$. who is emma stone father  \\
Before: \textcolor{blue}{Jeff Stone}, \textcolor{red}{Krista Stone} \\
After: \textcolor{blue}{Jeff Stone}  \\ \cline{1-1}
$5$. where did john steinbeck go to college  \\
Before:  \textcolor{red}{Salinas High School}, \textcolor{blue}{Stanford University}  \\
After: \textcolor{blue}{Stanford University} \\
\hline
\end{tabular}
\caption{Example questions and corresponding predicted answers before and after using unstructured inference. \textit{Before} uses (\textit{Structured~+~Joint}) model, and \textit{After} uses \textit{Structured~+~Joint~+~Unstructured} model for prediction. The colors \textsl{blue} and \textsl{red} indicate correct and wrong answers respectively.}
\label{tab:questions_list}
\vspace{-0.5cm}
\end{table}

\vspace{-0.2cm}
\subsubsection{Impact of the Inference on Unstructured Data}
\vspace{-0.1cm}
As shown in Table~\ref{tab:results}, when structured inference is augmented with the unstructured inference, we see an improvement of 2.9\% (from 44.1\% to 47.0\%). And when \textit{Structured~+~Joint} uses unstructured inference, the performance boosts by 6.2\% (from 47.1\% to 53.3\%) achieving a new state-of-the-art result. 
For the latter, we manually analyzed the cases in which unstructured inference helps. Table~\ref{tab:questions_list} lists some of these questions and the corresponding answers before and after the unstructured inference. We observed the unstructured inference mainly helps for two classes of questions: (1) questions involving aggregation operations (Questions~$1$-$3$); (2) questions involving sub-lexical compositionally (Questions~$4$-$5$). Questions~$1$ and~$2$ contain the predicate $largest$ an aggregation operator. A semantic parsing method should explicitly handle this predicate to trigger $max(.)$ operator. For Question $3$, structured inference predicts the Freebase relation \textsl{fb:teams..from} retrieving all the years in which \textsl{Ray Allen} has played basketball. Note that \textsl{Ray Allen} has joined \textit{Connecticut University}'s team in \textsl{1993} and \textsl{NBA} from \textsl{1996}. To answer this question a semantic parsing system would require a \textit{min}($\cdot$) operator along with an additional constraint that the year corresponds to the \textsl{NBA}'s term.  Interestingly, without having to explicitly model these complex predicates, the  unstructured inference helps in answering these questions more accurately. 
Questions~$4$-$5$ involve sub-lexical compositionally \cite{wang2015} predicates \textsl{father} and \textsl{college}. For example in Question~$5$, the user queries for the colleges that \textit{John Steinbeck} attended. However, Freebase defines the relation \textsl{fb:education..institution} to describe a person's educational information without discriminating the specific periods such as high school or college. Inference using unstructured data helps in alleviating these representational issues.

\vspace{-0.1cm}
\subsubsection{Error analysis}
\vspace{-0.1cm}
We analyze the errors of \textit{Structured~+~Joint~+~Unstructured} model. 
Around 15\% of the errors are caused by incorrect entity linking, and around 50\% of the errors are due to incorrect relation predictions. 
The errors in relation extraction are due to (i) insufficient context, e.g., in \textsl{what is duncan bannatyne}, neither the dependency path nor sentential context provides enough evidence for the MCCNN model; (ii) unbalanced distribution of relations~(3022 training examples for 461 relations) heavily influences the performance of MCCNN model towards frequently seen relations. The remaining errors are the failure of unstructured inference due to insufficient evidence in Wikipedia or misclassification.

\paragraph{Entity Linking.}
In the entity linking component, we had handcrafted POS tag patterns to identify entity mentions, e.g., DT-JJ-NN (noun phrase), NN-IN-NN (prepositional phrase). These patterns are designed to have high recall. 
Around 80\% of entity linking errors are due to incorrect entity prediction even when the correct mention span was found.

\paragraph{Question Decomposition.}
Around 136 questions (15\%) of dev data contains compositional questions, leading to 292 sub-questions (around 2.1~subquestions for a compositional question).
Since our question decomposition component is based on manual rules, one question of interest is how these rules perform on other datasets.
By human evaluation, we found these rules achieves 95\% on a more general but complex QA dataset QALD-5\footnote{\small\url{http://qald.sebastianwalter.org/index.php?q=5}}.

\subsubsection{Limitations}
While our unstructured inference alleviates representational issues to some extent, we still fail at modeling compositional questions such as \textsl{who is the mother of the father of prince william} involving multi-hop relations and the inter alia. Our current assumption that unstructured data could provide evidence for questions may work only for frequently typed queries or for popular domains like movies, politics and geography. We note these limitations and hope our result will foster further research in this area.

\section{Related Work}

Over time, the QA task has evolved into two main streams -- QA on unstructured data, and QA on structured data. TREC QA  evaluations \cite{voorhees1999trec} were a major boost to unstructured QA leading to  richer datasets and sophisticated methods \cite{wang2007jeopardy,heilman2010tree,yao2013answer,yih-EtAl:2013:ACL2013,yu2014deep,yang-yih-meek:2015:EMNLP,hermann2015teaching}. While initial progress on structured QA started with small toy domains like GeoQuery \cite{zelle1996learning}, recent focus has shifted to large scale structured KBs like Freebase, DBPedia \cite{unger2012template,DBLP:conf/acl/CaiY13,DBLP:conf/emnlp/BerantCFL13,DBLP:conf/emnlp/KwiatkowskiCAZ13,DBLP:conf/nlpcc/XuZFZ14}, and on noisy KBs \cite{banko2007open,carlson2010toward,krishnamurthy2012weakly,DBLP:conf/acl/FaderZE13,parikh:2015}. An exciting development in structured QA is to exploit multiple KBs  (with different schemas) at the same time to answer questions jointly \cite{yahya:2012,fader2014open,zhang2016joint}. QALD tasks and linked data  initiatives are contributing to this trend.

Our model combines the best of both worlds by inferring over structured and unstructured data. Though earlier methods exploited unstructured data for KB-QA \cite{krishnamurthy2012weakly,DBLP:conf/emnlp/BerantCFL13,yao-jacana-freebase-acl2014,reddy14,yih-EtAl:2015:ACL-IJCNLP}, these methods do not rely on unstructured data at test time. Our work is closely related to \newcite{joshi:2014} who aim to answer noisy telegraphic queries using both structured and unstructured data. Their work is limited in answering single relation queries. Our work also has similarities to \newcite{sun2015open} who does question answering on unstructured data but enrich it with Freebase, a reversal of our pipeline. Other line of very recent related work include \newcite{Yahya:2016:RQE:2835776.2835795} and \newcite{savenkovknowledge}.

Our work also intersects with relation extraction methods. While these methods aim to 
predict a relation between two entities in order to populate KBs \cite{mintz2009distant,hoffmann2011knowledge,DBLP:conf/naacl/RiedelYMM13}, we work with 
sentence level relation extraction for question answering. 
\newcite{krishnamurthy2012weakly} and \newcite{fader2014open}  
adopt open relation extraction methods for QA but they require hand-coded 
grammar for parsing queries. Closest to our extraction method is 
\newcite{yao-jacana-freebase-acl2014} and \newcite{yao-scratch-qa-naacl2015} who also uses sentence level relation extraction for QA. Unlike them, we can predict multiple relations per question, and our MCCNN architecture is more robust to unseen contexts compared to their logistic regression models. 

\newcite{dong-EtAl:2015:ACL-IJCNLP1} were the first to use MCCNN for question answering. Yet our approach is very different in spirit to theirs. Dong et al. aim to maximize the similarity between the distributed representation of a question and its answer entities, whereas our network aims to predict Freebase relations. Our search space is several times smaller than theirs since we do not require potential answer entities beforehand (the number of relations is much smaller than the number of entities in Freebase). In addition, our method can explicitly handle compositional questions involving multiple relations, whereas Dong et al. learn latent representation of relation joins which is difficult to comprehend. Moreover, we outperform their method by 7~points even without unstructured inference.

\section{Conclusion and Future Work}
We have presented a method that could infer both on structured and unstructured data to answer natural language questions. Our experiments reveal that unstructured inference helps in mitigating representational issues in structured inference. 
We have also introduced a relation extraction method using MCCNN which is capable of exploiting syntax in addition to sentential features. Our main model which uses joint entity linking and relation extraction along with unstructured inference achieves the state-of-the-art results on \webq dataset. A potential application of our method is to improve KB-question answering using the documents retrieved by a search engine.

Since we pipeline structured inference first and then unstructured inference, our method is limited by the coverage of Freebase. Our future work involves exploring other alternatives such as treating structured and unstructured data as two independent resources in order to overcome the knowledge gaps in either of the two resources. 

\section*{Acknowledgments}
We would like to thank Weiwei Sun, Liwei Chen, and the anonymous reviewers for their helpful feedback. This work is supported by National High Technology R\&D Program of China (Grant No. 2015AA015403, 2014AA015102), Natural Science Foundation of China (Grant No. 61202233, 61272344, 61370055) and the joint project with IBM Research. For any correspondence, please contact Yansong Feng.

\section*{Appendix}
\begin{figure}
 \centering
 \includegraphics[width=\columnwidth]{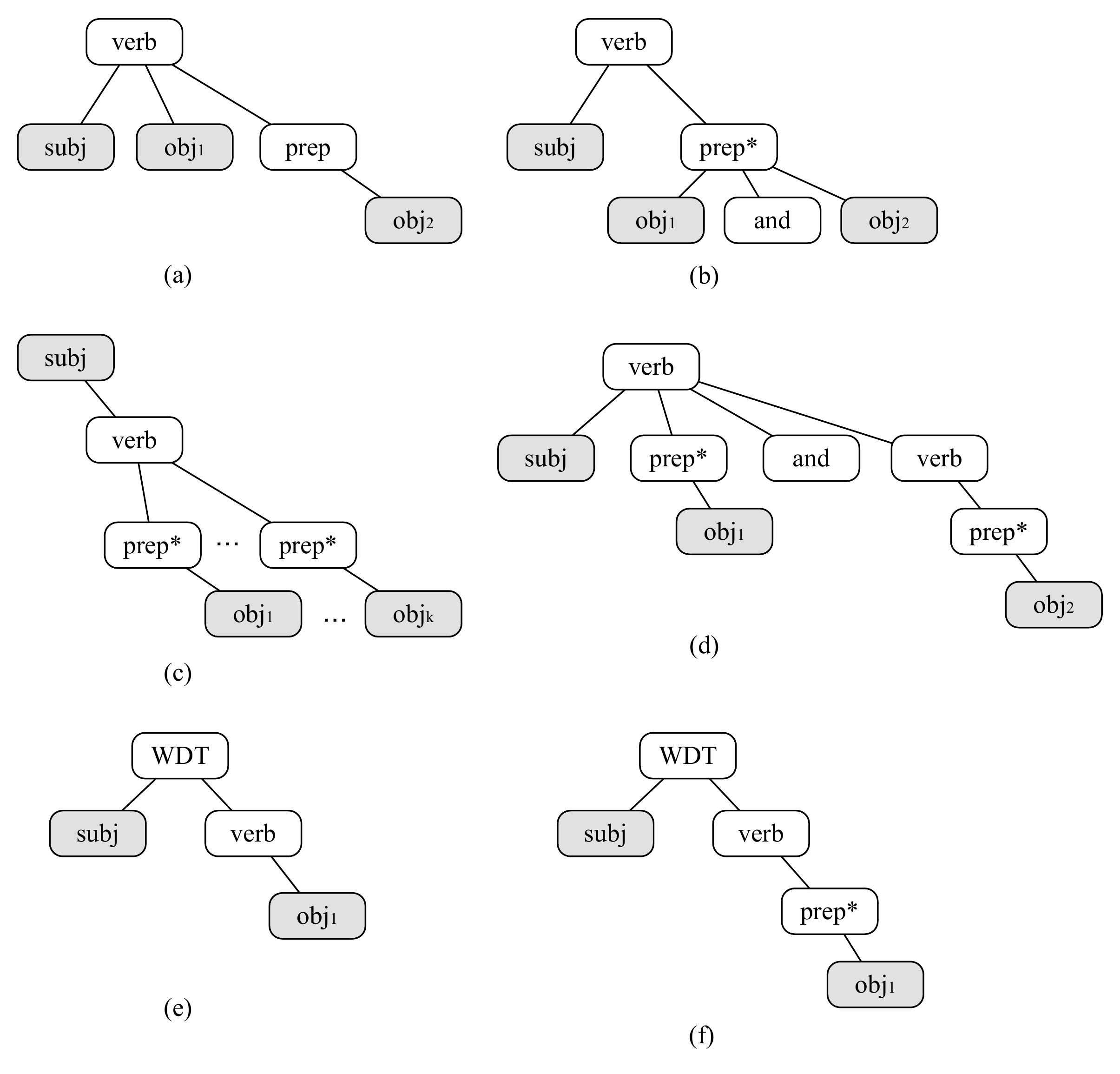}
 \caption{Syntax-based patterns for question decomposition.}
 \label{fig:patterns}
\end{figure}

The syntax-based patterns for question decomposition are shown
in \Cref{fig:patterns}. The first four patterns are designed to extract sub-questions from simple questions, while the latter two are designed for complex questions involving clauses.

\bibliography{naaclhlt2016}
\bibliographystyle{acl2016}

\end{document}